\documentclass[10pt,twocolumn,letterpaper]{article}

\usepackage[pagenumbers]{wacv} % To force page numbers, e.g. for an arXiv version

\usepackage{graphicx}
\usepackage{amsmath}
\usepackage{amssymb}
\usepackage{booktabs}
\usepackage{multirow}
\usepackage{algorithm}
\usepackage{algpseudocode}
\usepackage{enumitem}
\usepackage{soul}
\usepackage{pifont}% http://ctan.org/pkg/pifont
\newcommand{\xmark}{\ding{55}}%

\usepackage[pagebackref,breaklinks,colorlinks]{hyperref}

\usepackage[capitalize]{cleveref}
\Crefname{section}{Section}{Sections}
\Crefname{table}{Table}{Tables}
\Crefname{figure}{Figure}{Figures}
\newcommand{\dataset}{GazeSearch\xspace}
\newcommand{\method}{ChestSearch\xspace}

\def\wacvPaperID{1768} % *** Enter the WACV Paper ID here
\def\confName{WACV}
\def\confYear{2025}

\begin{document}

%%%%%%%%% TITLE - PLEASE UPDATE
\title{GazeSearch: Radiology Findings Search Benchmark}

\author{Trong Thang Pham$^{\ast}$, Tien-Phat Nguyen$^{\dagger}$, Yuki Ikebe$^{\ast}$, Akash Awasthi$^{\ddagger}$, Zhigang Deng$^{\ddagger}$,\\
Carol C. Wu$^{\mathsection}$, Hien Nguyen$^{\ddagger}$, and Ngan Le$^{\ast}$ \\
$^{\ast}$University of Arkansas, Fayetteville, AR, USA \\
$^{\dagger}$University of Science, VNU-HCM, Ho Chi Minh City, Vietnam \\
$^{\ddagger}$University of Houston, Houston, TX, USA \\
$^{\mathsection}$MD Anderson Cancer Center, Houston, TX, USA \\
% {\tt\small tp030@uark.edu, ntphat@selab.hcmus.edu.vn, yikebe@uark.edu, akashcseklu123@gmail.com,} \\
% {\tt\small zdeng4@central.uh.edu, CCWu1@mdanderson.org, hvnguy35@central.uh.edu, thile@uark.edu}
}
\maketitle

%%%%%%%%% ABSTRACT
\begin{abstract}
Medical eye-tracking data is an important information source for understanding how radiologists visually interpret medical images. This information not only improves the accuracy of deep learning models for X-ray analysis but also their interpretability, enhancing transparency in decision-making.
However, the current eye-tracking data is dispersed, unprocessed, and ambiguous, making it difficult to derive meaningful insights. Therefore, there is a need to create a new dataset with more focus and purposeful eye-tracking data, improving its utility for diagnostic applications.
In this work, we propose a refinement method inspired by the target-present visual search challenge: there is a specific finding and fixations are guided to locate it. After refining the existing eye-tracking datasets, we transform them into a curated visual search dataset, called \dataset, specifically for radiology findings, where each fixation sequence is purposefully aligned to the task of locating a particular finding.
Subsequently, we introduce a scan path prediction baseline, called ChestSearch, specifically tailored to \dataset. Finally, we employ the newly introduced \dataset as a benchmark to evaluate the performance of current state-of-the-art methods, offering a comprehensive assessment for visual search in the medical imaging domain. Code is available at \url{https://github.com/UARK-AICV/GazeSearch}.

\end{abstract}

%%%%%%%%% BODY TEXT
% total: 8 
% intro: 1 page 
% related work: 0.5 page 
% method: 3.5 page 
% experiment: 2.5 page 
% conclusion 0.5 page

\section{Introduction}
\label{sec:intro}

% ---- big picture
\begin{figure}[t]
    \centering
    \includegraphics[width=\linewidth]{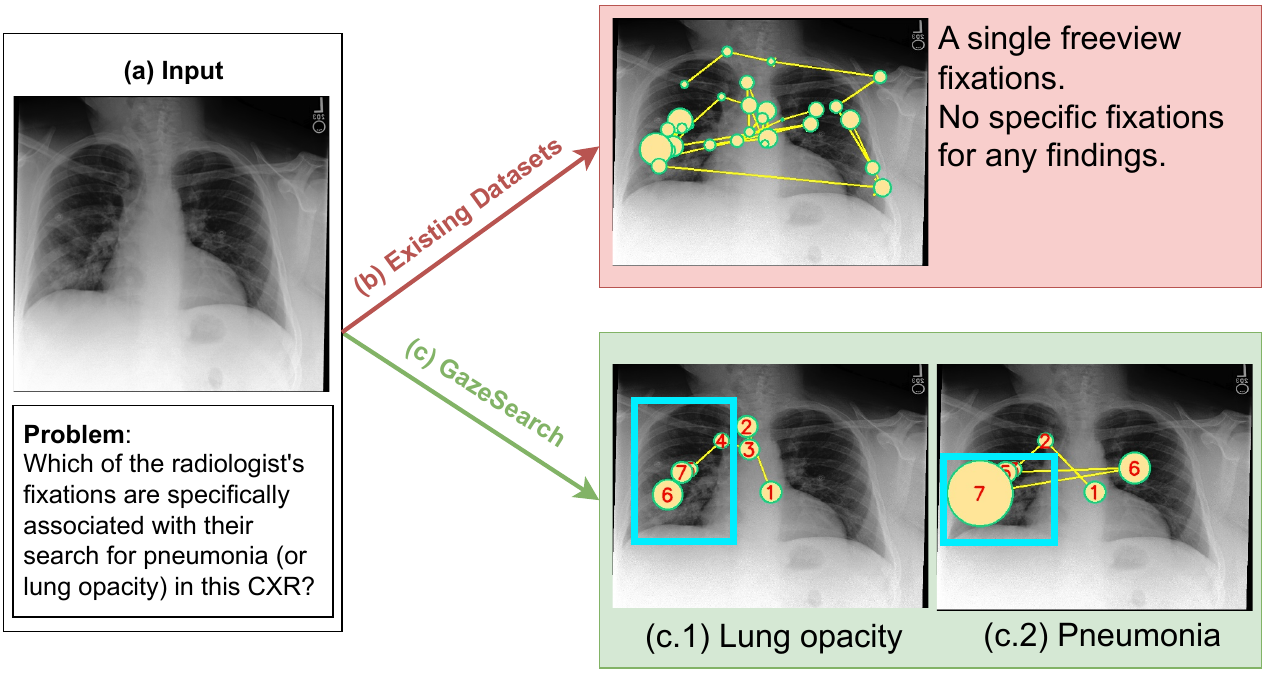}
\caption{(a) Given a CXR image, we are interested in radiologist's eye movement of radiologist when they search for a finding. (b) But, the existing eye gaze datasets are recorded in a free-view form, where fixations are distributed across the entire CXR image and making it unclear which fixations correspond to specific findings. (c) Our new GazeSearch dataset, where fixation sequence is focused for a specific finding. For example, the gaze sequence in (c.1) targets lung opacity, while (c.2) focuses on pneumonia. Each circle depicts a fixation, with the number and radius indicating its order and duration, respectively.}
\vspace{-1.2em}
    \label{fig:introduction}
\end{figure}

% Artificial Intelligence (AI) has been growing rapidly and become an important part of daily life~\cite{wu2019deep,waite2020review,houssami2019artificial,kasneci2023chatgpt,benzekry2020artificial,rubin2019artificial}, including important workers like clinical experts and healthcare providers. 
Artificial Intelligence (AI) has been growing rapidly and become an important part of daily life~\cite{baydoun2024artificial, yamazaki2022spiking, le2022deep, mir2023application, shang2024impact, yuksel2023review,nguyentruong2024vision,nguyen2024instance,le2024infty,le2024maskdiff,nguyen2024contrastive,khaldi2024unsupervised,nguyen2024tackling}, including important workers like clinical experts and healthcare providers~\cite{wu2019deep,waite2020review,houssami2019artificial,kasneci2023chatgpt,benzekry2020artificial,rubin2019artificial,le2021enhance}. Beyond achieving high performance, it is essential to develop AI systems that offer explainable and interpretable decision-making \cite{awasthi2024bridging, 10650009, vo2023aoe, yamazaki2023vltint, vo2022contextual, frasca2024explainable, hassija2024interpreting, pham2024ai,pham2024fgcxrradiologistalignedgazedataset}. This is especially important in sensitive domains such as healthcare, where credibility and reliability are critical to ensuring trust and safe implementation. Even though human experts remain the ultimate authority in decision-making, researchers are focusing on improving AI-assisted systems to reduce the burden for the experts. For example, we can use AI to produce preliminary results and the experts can either confirm or adjust~\cite{wu2019deep}. As a result, the collaborative approach between AI and professionals has successfully improved radiological diagnosis in many cases compared to radiologists or the system alone~\cite{waite2020review}. However, a key challenge is building trust in AI, especially with black-box models in healthcare, such as CXR analysis. This has increased the demand for models that mimic radiologists' behavior to improve interpretability. For instance, aligning AI systems with radiologists' visual attention patterns is essential~\cite{rudin2019stop,pham2024ai}. This has opened a new domain of research focused on modeling the radiologists' eye movements to improve the transparency and reliability of AI systems in clinical practice~\cite{castner2024expert}. 

Recognizing the importance of understanding how radiologists' eye movements impact diagnosis, datasets like EGD~\cite{karargyris2021creation} and REFLACX~\cite{bigolin2022reflacx} have been introduced. But, these eye-tracking datasets present two major challenges:

\noindent
\textit{Challenge \#1}: Free-view format - Existing eye-tracking datasets are collected in a free-view format, where fixations are distributed across the entire CXR image, making it unclear which fixations correspond to specific findings (as shown in \cref{fig:introduction} (b)). Moreover, these datasets often contain ambiguity and suffer from misalignment between the recorded fixations and the findings in the report, rendering them unsuitable for accurate scan path prediction.

\noindent
\textit{Challenge \#2}: Lack of finding-aware radiologist's scanpath models - Most existing scanpath prediction models~\cite{sounak:2023:gazeformer,zhibo:2022:targetabsent,zhibo:2020:cocosearch} are designed for general applications and lack the domain-specific expertise needed for radiology. Furthermore, current models trained on medical eye-tracking data are not tailored to the challenges of finding-aware visual search in radiology. For instance, I-AI~\cite{pham2024ai} only associates diseases with abnormalities in specific anatomical areas. While RGRG~\cite{tanida2023rgrg} uses anatomical bounding boxes without considering gaze for report generation.

To address the challenge \#1, we propose a finding-aware radiologist's visual search dataset, named \textbf{GazeSearch}. Our objective is to minimize the misalignment between the findings extracted from the radiology reports and their corresponding fixations. Insprired by the visual search datasets like COCO-Search18~\cite{zhibo:2020:cocosearch} or Air-D~\cite{shi:2020:air}, we further process GazeSearch by reducing the fixation length using a radius-based filtering heuristic, ensuring that the direction of fixations remains clear and manageable. Additionally, for every finding, we ensure that the duration of fixations within the location of the given finding is maximized. To create GazeSearch dataset, we utilize the existing free-view eye gaze datasets EGD~\cite{karargyris2021creation} and REFLACX~\cite{bigolin2022reflacx} (\cref{fig:introduction}(b)) to conduct a finding-aware radiologist's visual search dataset (\cref{fig:introduction}(c)), which produces two scanpaths for particular findings e.g., ``lung opacity'' (\cref{fig:introduction} (c.1)) and ``pneumonia''(\cref{fig:introduction} (c.2)) in this example. The goal of releasing this dataset is to foster the development of algorithms that better mimic radiologists, especially focusing on understanding observation sequences, attention (duration), frequency on key regions, and expert knowledge~\cite{wu2021chest,neves2024shedding}.

To address challenge \#2, we introduce \textbf{\method}, a scanpath prediction architecture that surpasses existing models. \method builds on a standard meta architecture~\cite{cheng2021per} featuring a feature extractor~\cite{he2015resnet,lin2017fpn} and a Transformer decoder~\cite{vaswani2017attention}, with two key enhancements. First, we train the feature extractor using the self-supervised MGCA method~\cite{wang2022mgca} on the large MIMIC-CXR~\cite{johnson2019mimic} dataset, providing a strong initialization for training. Second, we utilize the modified cross attention from \cite{cheng2022masked} with a query mechanism to select only relevant fixations for predicting the next fixation. Then, the model's three heads handle distinct tasks: predicting 2D coordinates, duration, and stopping points. Finally, we benchmark \method against current state-of-the-art visual search models on \dataset, showcasing the current advancements in radiology visual search.

% leverages many domain-specific techniques. Specifically, we create a pretrained checkpoint on the MIMIC-CXR~\cite{johnson2019mimic} dataset as our feature extractor. In the transformer decoder, we design the prediction process to leverage previous fixations for predicting a single next fixation, including the 2D coordinate and the duration. Given that the number of findings is typically fixed and adheres to the Chexpert~\cite{irvin2019chexpert} format, which has 13 abnormal findings, we use a query mechanism to let the model flexibly select the most relevant fixation embedding before making predictions. We then benchmark the proposed method with various state-of-the-art scanpath prediction methods on our curated dataset to demonstrate the current state of the medical visual search task.  

%---- summary contribution 

Our main contributions are:
\begin{itemize}[noitemsep,topsep=0pt]
    \item \textbf{\dataset:} We propose a processing technique that converts free-view eye gaze data into finding-aware radiologist's visual search data. This curated dataset is the first target-present visual search dataset for chest X-ray, making possible deep learning modeling of medical visual search prediction.
    \item \textbf{\method:} We propose a transformer-based model that utilizes a radiology pretrained feature extractor and query mechanism to choose only relevant fixations to predict subsequent fixations based on previous ones. Additionally, we evaluate \method against several leading generic scanpath prediction models using our \dataset to showcase the current progress in the medical visual search task.
\end{itemize}

\section{Related works}
\label{sec:related-work}
\noindent 
\textbf{Visual Search Datasets.} Search datasets have been rising recently due to the interest in understanding human behavior~\cite{jiang2015salicon,papadopoulos2014training,ehinger2009modelling,zelinsky2019benchmarking,gilani2015pet,shi:2020:air,shuo:2015:austim}. This is particularly evident in the general visual domain, where numerous datasets have been created across diverse settings. These datasets cover a wide range of scenarios, from searching for multiple targets simultaneously~\cite{gilani2015pet} to focusing on a single or two target categories~\cite{ehinger2009modelling,zelinsky2019benchmarking}. Some datasets, like COCO-Search18~\cite{zhibo:2020:cocosearch}, feature a large number of target objects, or adopt a Visual Question Answering approach~\cite{shi:2020:air}.
In contrast, the medical domain has lagged behind in terms of dedicated visual search datasets. Existing medical datasets primarily focus on multi-target search tasks, as demonstrated by datasets like EGD~\cite{karargyris2021creation} and REFLACX~\cite{bigolin2022reflacx}. However, there is a significant lack of search datasets tailored for the medical domain. 
This paper makes a novel contribution by addressing this research gap. We introduce the first target-present visual search dataset specifically designed for the medical field. This dataset opens up new avenues for research and development in this critical area.

\noindent 
\textbf{Visual Search Baselines.} Parallel to the growth of visual search datasets, significant advancements have been made in scan path prediction accuracy~\cite{zhang2018finding,adeli2018deep,Wei-etal-NIPS16,matthias:2016:deepgaze,marcella:2018:sam}. Early scanpath models mostly rely on sampling fixations from saliency maps~\cite{wei:2011:stde,olivier:2015:saccadicmodel,calden:2018:star-fc,laurent:1998:visualattention}. 
Recent advancements, including the integration of deep neural networks~\cite{xianyu:2021:vqa,sounak:2023:gazeformer,wanjie:2019:iorroi,zhibo:2020:cocosearch,zhibo:2023:humanattention,zhibo:2022:targetabsent}, reinforcement learning techniques~\cite{xianyu:2021:vqa,zhibo:2020:cocosearch,zhibo:2022:targetabsent}, and transformer-based architectures~\cite{mengyu:2023:scanpath,sounak:2023:gazeformer,yang2024hat,chen2024isp}, have significantly deepened our understanding of the temporal dynamics of human attention.  However, generic models are designed for broad application, so the performance of generic visual search models on CXR is uncertain and potentially subpar. This work introduces a transformer-based method that can work well without these restrictive assumptions. Additionally, we further conduct a comparative experiment between state-of-the-art methods from the general visual domain and our proposed method, providing a comprehensive evaluation of their performance in the medical domain.

\section{GazeSearch Dataset}
\label{sec:dataset}

\begin{figure}
    \centering
    \includegraphics[width=\linewidth]{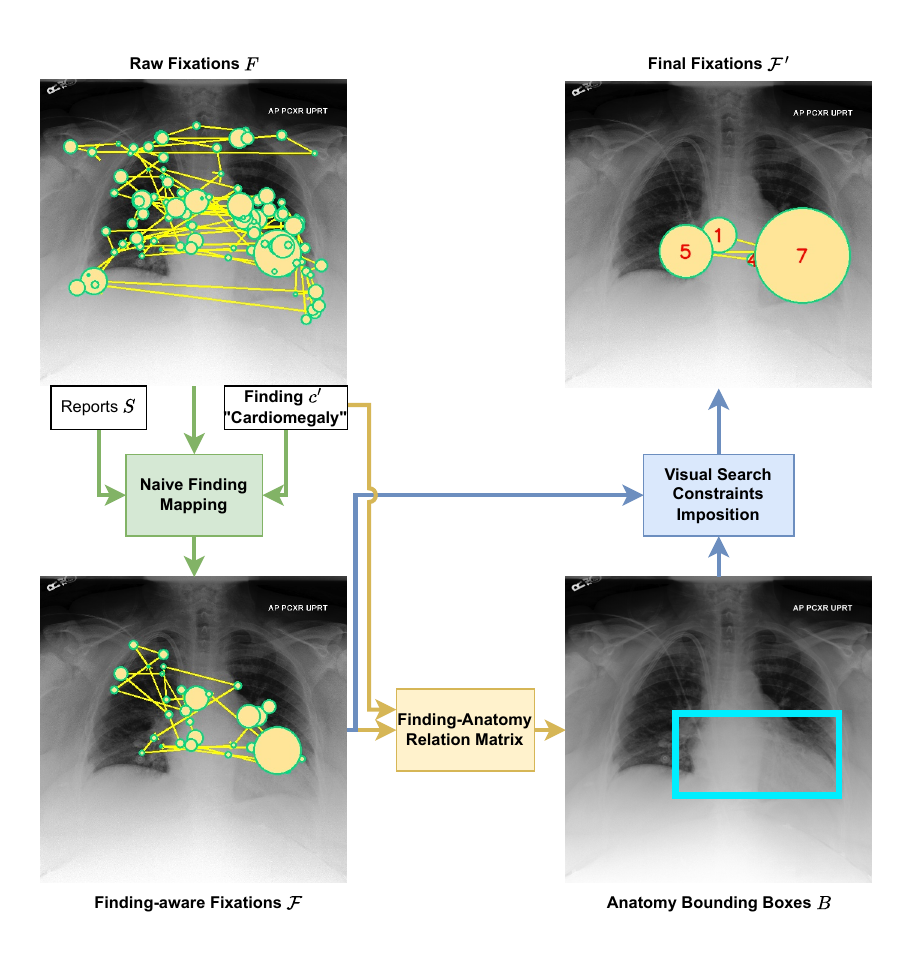}
    \caption{Pipeline of \dataset creation, which processes free-view eye gaze data as input and outputs a finding-aware scanpath.}
    % https://drive.google.com/file/d/139SP0EvXt_29gJWEhiwMiB_5yU03USz4/view?usp=sharing
    \label{fig:data-processing}
\end{figure}

\begin{algorithm}[!t]
\caption{Radius-based Filtering Procedure}\label{alg:radius-filtering}
\begin{algorithmic}
\State \textbf{Input:} Image width $W$, image height $H$, bounding boxes $B$, max length $M$, radius $r$, fixations \(\mathcal{F} = \{(x_1, y_1, d_1), (x_2, y_2, d_2), \dots, (x_n, y_n, d_n)\}\)
\State \textbf{Output:} Filtered fixations \(\mathcal{\hat{F}}\)
\State \textbf{Initialize:} \(\mathcal{\hat{F}} = (W/2, H/2, 0.3)\) 
\State // The last point must be inside $B$.
\State $j \gets \max \{i| (x_i,y_i) \in B,  (x_i,y_i, d_i) \in \mathcal{F}, 1 \leq i \leq n\}$

\State // Apply radius heuristic with looping backward.
\State $c \gets \{(x_j, y_j)\}$, where $(x_j, y_j, d_j) \in \mathcal{F}$
\For{each point \( (x_i, y_i,d_i) \in \mathcal{F} \) from $j-1$ to 1}
    \If{\( (x_{i}, y_{i}) \) is within radius $r$ of \( (x_{i+1}, y_{i+1}) \)}
        \State $c \gets c \cup \{(x_{i}, y_{i})\}$
    \Else
        \State $x \gets \frac{1}{|c|} \sum_k x_k, y \gets \frac{1}{|c|} \sum_k y_k, d \gets \sum_k d_k$, 
        \State where $(x_k,y_k,d_k) \in c$
        \State $c \gets \{(x_{i}, y_{i})\}$
        \State $\mathcal{\hat{F}} \gets \mathcal{\hat{F}} \cup \{(x,y,d)\}$
        \If{$|\mathcal{\hat{F}}| = M$}
            \State \textbf{break}
        \EndIf
    \EndIf
\EndFor
\If{$c \neq \{\}$ \textbf{and} $|\mathcal{\hat{F}}| < M$}
    \State $x \gets \frac{1}{|c|} \sum_k x_k, y \gets \frac{1}{|c|} \sum_k y_k, d \gets \sum_k d_k$, 
    \State where $(x_k,y_k,d_k) \in c$
    \State $\mathcal{\hat{F}} = \mathcal{\hat{F}} \cup \{(x,y,d)\}$
\EndIf  
\end{algorithmic}
\end{algorithm}

\begin{algorithm}[!t]
\caption{Time-spent Constraining Procedure}\label{alg:time-spent-constraint}
\begin{algorithmic}
\State \textbf{Input:} \(\mathcal{\hat{F}} = \{(x_1, y_1, d_1), (x_2, y_2, d_2), \dots, (x_n, y_n, d_n)\}\), bounding boxes $B$
\State \textbf{Output:} Constrained fixations \(\mathcal{F'}\)
\State $d^{out} \gets \{\sum_{i=k, (x_i,y_i)\notin B}^{n} d_i | (x_k,y_k,d_k) \in \mathcal{\hat{F}}, 1 \leq k \leq n \}$.
\State $d^{in} \gets \{\sum_{i=k,(x_i,y_i)\in B}^{n} d_i | (x_k,y_k,d_k) \in \mathcal{\hat{F}},  1 \leq k \leq n \}$.
\State $D \gets \{i | d_i^{in} \geq d_i^{out}, 1 \leq i \leq n\}$.
\If{$1 \notin D$}
    \State $t \gets \min D$
    \State $\mathcal{F'} \gets \{(x_i,y_i,d_i) | i \geq t , (x_i,y_i,d_i) \in \mathcal{\hat{F}}\}$
\EndIf
\end{algorithmic}
\end{algorithm}

When studying free-view eye-tracking datasets from sources like REFLACX~\cite{bigolin2022reflacx} and EGD~\cite{karargyris2021creation}, we notice that the eye-tracking data (including both gaze and fixations) is often ambiguous and lacks clarity. This ambiguity comes from the data collection settings, where radiologists look for multiple findings simultaneously. As a result, each fixation captures visual information relevant to multiple findings rather than a specific finding. Therefore, the fixations from these eye-tracking datasets are unsuitable for studying their relationship to specific findings, i.e. addressing the visual search problem.
Additionally, when visualizing these gaze points or fixations over an image, they often cover more than 80\% of the lung area, even though the actual anomaly area might be much smaller. We calculate the fixation coverage distribution in Supplementary.
This raises a concern that using the free-view fixations from the given datasets may not be effective and could even pose risks in sensitive sectors like healthcare, particularly for tasks requiring precise localization of specific findings.

To solve this issue, one way is to collect eye-tracking data under the visual search setting directly. 
However, to collect data by having radiologists examine each of the 14 standard findings in CheXpert~\cite{irvin2019chexpert}, would be costly and time-consuming. 
Therefore, this paper will propose an alternative technique that leverages eye-tracking data directly from the free-view setting to convert to the finding-aware visual search setting.

Inspired by visual search, we studied the COCO-Search18~\cite{zhibo:2020:cocosearch}, Air-D~\cite{shi:2020:air}, and COCO-Freeview~\cite{chen2022cocofreeview1,zhibo:2023:humanattention}, and identified two key properties that are required in a visual search dataset:

\noindent
\textit{Property \#1}: Late fixations tend to converge to more decisive regions of interest (ROIs)~\cite{shi:2020:air}. And, Shi et al.\cite{shi:2020:air} have concluded the late fixations are for searching.

\noindent
\textit{Property \#2}: The fixations within the object of interest tend to have longer durations, while those outside the object are typically shorter.

Based on those two facts, we propose an approach to convert from free-view data into a visual search format, ensuring the filtered fixations retain properties \#1 and \#2 without sacrificing too many fixations. 
% While this approach cannot replace the real data captured directly from the visual search setting, we believe it can be a way to leverage the free-view datasets to study the visual search behavior of radiologists. 
\cref{fig:data-processing} illustrates the overview of our data processing pipeline, including Naive Finding Mapping (\cref{sec:naive-finding-mapping}) to clean irrelevant fixations for a given finding, Finding-Anatomy Relation Matrix (\cref{sec:finding-anatomy-relation-matrix}) to extract key regions, and finally Visual Search Constraint Imposition (\cref{sec:visual-search-constraint-imposition}) to produce the fixations that have both visual search properties.

\subsection{Naive Finding Mapping}
\label{sec:naive-finding-mapping}
The first problem we must solve is the mismatch between the fixations and the corresponding radiologists' report sentences. The main reason is radiologists observe the images first and then describe their findings, meaning the fixations within the time frame of a sentence may not fully capture the findings reported.
Inspired by I-AI~\cite{pham2024ai}, we start by completely removing fixations after the current spoken sentence.  Let $S = \{s_1, s_2, ..., s_{|S|}\}$ be the sequence of sentences in the transcript. 
Let $C = \{c_1, c_2, ..., c_m\}$ be the set of possible findings (e.g., CheXpert labels). We define a function $\phi: S \rightarrow C $ where $c_j = \phi(s_i)$ if sentence $s_i$ corresponds to finding $c_j$. 
In our implementation, $\phi(\cdot)$ is the Chexbert model~\cite{smit2020chexbert}.
For a target finding $c' \in C$, let $u = \max \{i | \phi(s_i) = c', 1 \leq i \leq |S|\} $. 
Then, the new finding-aware fixations $\mathcal{F}$ for $c'$ is 
\begin{equation}
    \mathcal{F} = \{ (x_i, y_i, t_i, d_i) | (x_i, y_i, t_i, d_i) \in F, 0 \leq t_i \leq e_u \}
\end{equation}
where $F = \{(x_1, y_1, t_1, d_1),.., (x_{|F|}, y_{|F|}, t_{|F|}, d_{|F|})\}$ is the free-view fixations, with $(x_i, y_i)$ as spatial coordinates, $t_i$ as captured timestamp, and $d_i$ as duration, and $e_u$ is the ending time of the sentence $s_u$. 
From this point onwards, we only use the triplet $(x_i, y_i, d_i)$ and ignore the captured timestamp $t_i$ for our fixation sequence: $\mathcal{F} = \{(x_1,y_1,d_1), \dots, (x_n,y_n,d_n)\}$, where $n = |\mathcal{F}|$ is the fixation sequence length.

\subsection{Finding-Anatomy Relation Matrix}
\label{sec:finding-anatomy-relation-matrix}
% Ideally, if we had a highly accurate radiology finding detector, it would allow us to identify the key regions associated with specific findings. However, to the best of our knowledge, a reliable detector that can predict bounding boxes $B$ for particular findings has not yet been developed.

To address this, we leverage the Chest ImaGenome~\cite{wu2021chest} dataset, which offers pairs of findings and their corresponding anatomies, along with anatomy bounding boxes linked to each finding. For precision, we rely on the gold subset of Chest ImaGenome to construct a relation matrix between findings and anatomies. As a final step, a radiologist with over 15 years of experience thoroughly reviews and refines the matrix. The finalized matrix is included in the Supplementary Material.
Once the relation matrix is completed, we reference the given finding $c'$ to identify the corresponding anatomies and utilize the ground truth anatomy bounding boxes provided by Chest ImaGenome as our $B$ for the subsequent steps.

\subsection{Visual Search Constraint Imposition}
\label{sec:visual-search-constraint-imposition}
After \cref{sec:naive-finding-mapping}, the maximum fixation sequence length can be over 340 fixations for a finding. Therefore, another task we must solve is reducing this length to an interpretable level for humans.

Utilizing both properties (1) and (2) as our guidance for this process, we perform two main steps: radius-based filtering (to enforce property \#1) and time-spent constraining (to enforce property \#2). 
Besides property \#1, we observe that the captured fixations from EGD and REFLACX cover one-degree visual angle~\cite{lemeur2013methodscomparingscanpaths,karargyris2021creation,bigolin2022reflacx}. Based on that fact, we use the \cref{alg:radius-filtering} to cluster the finding-aware fixations $\mathcal{F}$ to create another fixation set $\mathcal{\hat{F}}$, with a larger radius $r$ of two-degree of visual angle and $M$ is the max length of fixation sequence. Property \#1 is enforced by iterating backward from the end to the beginning of the fixation sequence $\mathcal{F}$. 
Then, we use the \cref{alg:time-spent-constraint} to make sure the late fixations must spend the most time in the anatomies of interest, which satisfies property \#2.

In \cref{alg:radius-filtering,alg:time-spent-constraint}, we define a point $(x,y)$ to be in the bounding box sets $B$ for notation convenience:
\begin{align}
    (x,y) \in B \iff & x^{left} \leq x \leq x^{right}, y^{top} \leq y \leq y^{bottom}, \notag \\ & \forall (x^{left},y^{top},x^{right},y^{bottom}) \in B
\end{align}
To align with the COCO-Search18 dataset, we set the maximum fixation length to $M = 7$ and add a default center as the start fixation. This choice is based on the observation that 95\% of the samples in COCO-Search18 have fixation lengths under 7. For the first fixation's duration, we assign 0.3 seconds to it, which reflects the duration of 91\% of first fixations in COCO-Search18.
In total, \dataset has 2,081 images with 413 samples from EGD and 1,668 samples from REFLACX. There are a total of 13 findings. Each sample has fixations for 1 to 6 findings and has a max length of 7, including the default middle fixation. For training and evaluation, we split the dataset into 1,456 samples for training (70\%), 208 samples for validation (10\%), and 417 samples for testing (20\%).

\begin{table}[!t]
\centering
\caption{Usage validation experiments on our \dataset. mHC (mean Heatmap Coverage) is the average ratio of the heatmap area to the lung area across all images in \dataset.}
\label{tab:usage-validation}
\resizebox{\linewidth}{!}{
\begin{tabular}{l|l|c| c}
\toprule
\textbf{Method}                   & \textbf{Fixation Type} & \textbf{AUC} & \textbf{mHC}\\ \hline 
Naive Classifier & \xmark & 76\% & \xmark \\ \hline 
\multirow{2}{*}{Temporal Classifier} & Freeview      & 81\% &  91\% \\ 
                          & \textbf{Finding-aware (\dataset)} & \textbf{81\%} & \textbf{44\%} \\ \bottomrule

\end{tabular}
}
\vspace{-1.2em}
\end{table}

\subsection{Usage Validation}
Filtering fixations requires discarding information, so it is essential to test and ensure that the new data remains valuable.
To validate that \dataset's fixations can be as useful as the free-view fixation maps from EGD and REFLACX, we follow Karargyris et al.~\cite{karargyris2021creation} to perform the Temporal Heatmap experiment. This experiment evaluates whether eye gaze data can enhance classifier performance when using ground truth fixations as temporal inputs.
The results, \cref{tab:usage-validation}, indicate that despite using only half the area compared to the free-view setting, performance remains comparable. Detailed implementation of this experiment is provided in the Supplementary.

\begin{figure}[!t]
    \centering
    \includegraphics[width=\linewidth]{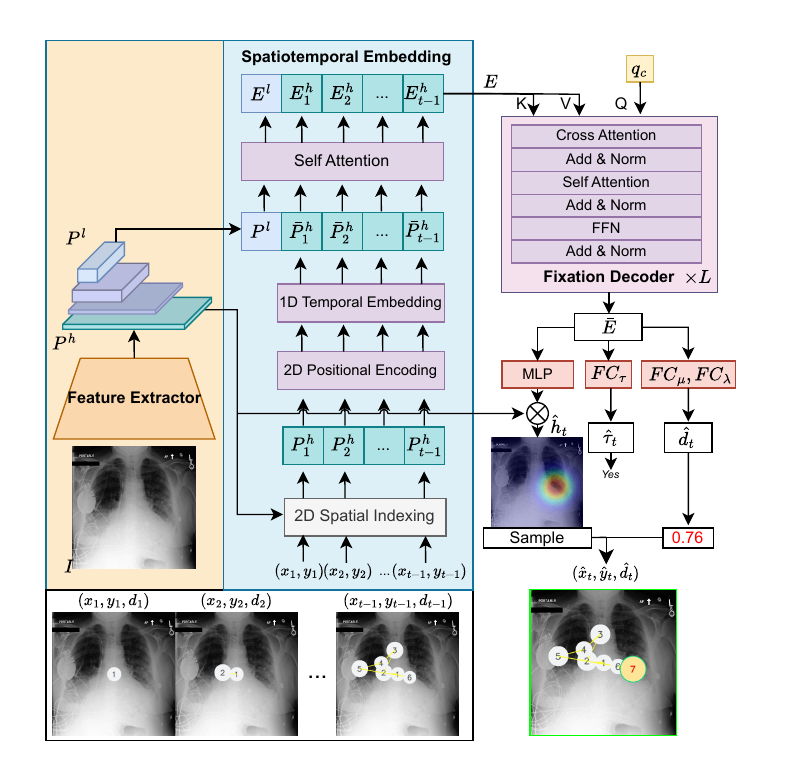}
    % https://drive.google.com/file/d/1deGksUfFB01HPAQUMxlihaKqWjoRuez5/view?usp=sharing
    \caption{The figure provides a detailed view of \method. It begins by processing the previous fixations, denoted as \( \{(x_i, y_i, d_i)\}_{i=1}^{t-1} \), along with the input chest X-ray image \( I \), through a Feature Extractor and Spatiotemporal Embedding to generates the spatiotemporal embedded feature \( E \). Next, the Fixation Decoder uses a learnable query \( q_c \) and the embedded feature \( E \) to decode it into a feature \( \bar{E} \). From here, three heads use \( \bar{E} \) to predict the next fixation coordinates \( (\hat{x}_t, \hat{y}_t, \hat{d}_t) \). Here, at step \( t \), the termination head outputs ``Yes,'' indicating that this is the final fixation for the image \( I \).}
    \vspace{-1em}
    \label{fig:proposed-method}
\end{figure}

% \begin{figure}[!t]
%     \centering
%     \includegraphics[width=\linewidth]{imgs/detail.pdf}
%     % https://drive.google.com/file/d/1deGksUfFB01HPAQUMxlihaKqWjoRuez5/view?usp=sharing
%     \caption{The detailed illustration of our proposed method. }
%     \label{fig:proposed-method}
% \end{figure}
\section{\method}
Given a CXR image $I$ of dimensions $H \times W$ and a target finding $c'$, our objective is to generate a scan-path comprises of fixations $y=\{f_i\}_{i=1}^n$, where $n$ represents the number of fixations, and $f_i = (x_i, y_i, d_i)$ is the fixation at 2D coordinate $(x_i, y_i)$ with a duration of $d_i$.  

\cref{fig:proposed-method} provides an overview of our method. The process begins by applying a Feature Extractor (\cref{sec:feature_extractor}) to process $I$ to extract both detailed and high-level visual features. Following this, a Spatiotemporal Embedding (\cref{sec:embedding}) embeds previous fixations, combined with multi-resolution features, to capture contextual relationships within the sequence. These features are passed through a transformer decoder with cross-attention, self-attention, feedforward layers, and normalization (\cref{sec:fixation_decoder}) to create a decoded latent feature. Finally, the decoded feature is fed into three heads to predict the next fixation: termination prediction (\cref{sec:termination-head}), fixation duration (\cref{sec:duration-head}), and distribution for the next fixation (\cref{sec:fixation-distribution-head})

\subsection{Feature Extractor}
\label{sec:feature_extractor}
Using features from only the last layer is inadequate for predicting scanpaths~\cite{zhibo:2022:targetabsent}. Therefore, we employ ResNet-50 FPN~\cite{lin2017fpn} as our Feature Extractor module (FE). Besides, using the ImageNet~\cite{deng2009imagenet} checkpoint may not be optimal for the medical domain, so we train the FE using a self-supervised approach based on MGCA~\cite{wang2022mgca} with the MIMIC-CXR dataset~\cite{johnson2019mimic}.
% MGCA~\cite{wang2022mgca} trains ResNet-50 at both global and local (patch) levels, which aligns with our need to utilize patch features in subsequent steps.
From the CXR image $I$ with size \( H \times W \), FE produces four multi-scale feature maps \( P = \{P^1, \dots, P^4\} \). Then we need to mimic how human see an image: at first we only see the image at a high level understanding, with no clear details, and then we look carefully to search for what we need~\cite{zhibo:2020:cocosearch}. So we use one feature map with low resolution \( P^l = P^1 \in \mathbb{R}^{C \times \frac{H}{32} \times \frac{W}{32}} \), where \( C \) is the channel dimension, to represent high-level visual feature, and one high-resolution feature map \( P^h = P^4 \in \mathbb{R}^{C \times \frac{H}{4} \times \frac{W}{4}} \) to represent detailed visual information. 

\subsection{Spatiotemporal Embedding} 
\label{sec:embedding}
Given the previous predicted fixations $\{(x_i,y_i)\}_{i=1}^{t-1}$, $P^l$, and $P^h$, we then embed the previous fixations to create the feature list as the input for the decoder in \cref{sec:fixation_decoder}. 

\noindent
\textbf{2D Spatial Indexing.} Every $(x_i, y_i)$, where $0 \leq x_i \leq W$ and $0 \leq y_i \leq H$, is scaled down to the same resolution as of $P^h$, which result in the new $0\leq x'_i \leq \frac{W}{4}$ and $0 \leq y'_i \leq \frac{H}{4}$ in our case. Then, we index the feature cell at the coordinate $(x'_i, y'_i)$ in $P^h$, called $P_i^h$. We will have the list of feature $\{P_i^h\}_{i=1}^{t-1}$. 

\noindent
\textbf{2D Positional Embedding.} For every $P_i^h$, we encode the spatial information by using positional encoding twice, first in the x-axis, then in the y-axis. As the 2D order is important, we enforce the sinusoid version of positional encoding. 

\noindent
\textbf{1D Temporal Embedding.} We also need to let the model know the order of each fixations. However, the role of fixation order in diagnosing CXR in practice is complicated, so we let the model decide the embedding by applying a learnable position embedding here. This results in the $\{\bar{P}^h_i\}_{i=1}^{t-1}$ sequence of embedded feature.

\noindent
\textbf{Self Attention.} Finally, we feed $\{\bar{P}^h_i\}_{i=1}^{t-1}$ into several layers of self-attention to aaggregate information so that each position is influenced by the relevant fixations. In the self-attention layers, we also provide the model with a low-resolution feature map $P^l$ to supply high-level feature information. This intuition is also proven effected empirically, as it will be shown later in \cref{sec:ablation-study}. The final embeddings are $E = \{E^l\} \cup \{E^h_i\}_{i=1}^{t-1}$, where $E^l \in \mathbb{R}^{D \times \frac{H}{32} * \frac{W}{32}}$ and $E^h_i \in \mathbb{R}^{D}$. 

\subsection{Fixation Decoder}
\label{sec:fixation_decoder}
At this layer, we have the finding list $q  = \{q_c\}_{c}^{|q|}$ which serves as the set of queries. The number of queries is the number of findings in our dataset $|q| = 13$ with $q_c \in \mathbb{R}^{D}$ is a learnable embedding for the current finding query $c$. 
The previous module (\cref{sec:embedding}) gives us the embeddings of previous fixations $E$.

The Fixation Decoder module is the modified transformer decoder~\cite{cheng2022masked} including the blocks as shown in \cref{fig:proposed-method}. The cross-attention block uses the query embedding $q$ as the query input Q, with $E$ serving as both key (K) and value (V). This allows the model to capture the correlations among previous fixations and accurately predict the next fixation. The resulting feature then passes through self-attention layers, residual connections, normalization, and a feed-forward network. This process repeats for $L$ layers in the decoder. The final output $\bar{E} \in \mathbb{R}^{|q| \times D}$ is then processed by three different heads.

\subsection{Termination Head}
\label{sec:termination-head}
A fixation sequence's length can vary, so our model needs to learn when to stop.
 To achieve this, we use a head consisting of a fully connected (FC) layer followed by a sigmoid function that maps $\bar{E}$ to termination value i.e., $\hat{\tau}_{t}^c \in \mathbb{R} = \text{sigmoid}(FC_{\tau} (\bar{E}))$.

\subsection{Duration Head}
\label{sec:duration-head}
The duration can be considered as a Gaussian distribution. We use $\bar{E}$, then regress it into a mean value $\mu_{d_t} = FC_\mu(\bar{E})$ and a log-variance $\lambda_{d_t} = FC_\lambda(\bar{E})$:
\begin{equation}
\begin{aligned}
\hat{d}_t & = \mu_{d_t} + \epsilon_{d_t} \cdot \exp(0.5 \lambda_{d_t}), \\
\epsilon_{d_t} & \sim \mathcal{N}(0, 1) \\
\end{aligned}
\end{equation}
where $\epsilon_{d_t}$ noise gives our prediction a probabilistic characteristic, and $\hat{d}_t \in \mathbb{R}^{|q|}$ is the duration prediction.
The inspiration comes from using the reparameterization trick~\cite{doersch2016tutorial}, which allows us to backpropagate from the label back to the normal distribution.

\subsection{Distribution Head}
\label{sec:fixation-distribution-head}
Because fixation is random in nature, we predict a 2D distribution in the form of a heatmap $\hat{h}_t \in [0,1]^{|q| \times (\frac{H}{4} * \frac{W}{4})}$. Formally, we compute: 
\begin{align}
    \bar{E}' &= \text{MLP}(\bar{E}) \notag \\ 
    \hat{h_t} &= \text{sigmoid}(\text{Matmul}(\bar{E}', P^h))
\end{align}
where Matmul$(\cdot, \cdot)$ is the matrix multiplication between two input tensors, and $\bar{E}' \in \mathbb{R}^{|q| \times D}$ is the latent embedding prepared for heatmap generation. At inference, we sample the next 2D coordinate $\hat{f}_{t}=(\hat{x}_{t},\hat{y}_{t})$ from the distribution map $\hat{h}_t$ for every given timestamp $t$.

\subsection{Objective Functions}
\label{sec:losses}
% Note that while the model can predict fixation distribution for all queries, shown in \cref{fig:proposed-method}, at computing losses we only pick the current task for the current sample. This will relax us from the constraint ``every image must have all $|q|$ query'', which is not realistic. So in the below equations, we only use one dimension from $|q|$ query.
\method has three objectives, each corresponding to one of its heads: the loss between the ground truth and predicted distributions, the loss for termination, and the loss for duration.

The termination loss is just a standard binary cross-entropy between the predicted termination value $\hat{\tau}_t$ and the corresponding ground truth $\tau_t$.

\begin{equation}
    \mathcal{L}_{\tau} = - \tau_t \log (\hat{\tau}_t) - (1-\tau_t)\log (1-\hat{\tau}_t), 
\end{equation}

The distribution loss is defined as focal pixel-wise loss:
\begin{equation}
\resizebox{0.90\columnwidth}{!}{$
\mathcal{L}_{h} = -\frac{1}{N} \sum_{ij} \left\{
\begin{array}{ll}
(1 - \hat{h}_{ij})^{\gamma} \log(\hat{h}_{ij}) & \text{if } h_{ij} = 1, \\
\begin{gathered}
(1 - h_{ij})^{\alpha} (\hat{h}_{ij})^{\gamma} \log(1 - \hat{h}_{ij})
\end{gathered} & \text{otherwise}, 
\end{array} \right.
$}
\label{eq:focal-pixel-loss}
\end{equation}
where $0\leq i \leq \frac{H}{4}$, $0\leq j \leq \frac{W}{4}$ are the 2D indexes, $N =  \frac{H}{4} *  \frac{W}{4}$ is the number of values, $\alpha$ and $\gamma$
are the hyper-parameters indicating the importance of each pixel. 
The duration loss is defined as the L1 loss, i.e., $\mathcal{L}_{d} = |\hat{d}_t - d_t|$.
% \begin{equation}
% \mathcal{L}_{d} = |\hat{d}_t - d_t|
% \end{equation}

Finally, we train all three losses jointly.
\begin{equation}
    \mathcal{L} = \mathcal{L}_{\tau} + \mathcal{L}_{h} + \mathcal{L}_{d}
    \vspace{-1em}
\end{equation}
%-------------------------------------------------------------------------
\begin{table*}[!t]
\centering
\caption{Performance comparison between our ChestSearch and SOTA visual search methods.} 
{
\footnotesize
\begin{tabular}{l|cc|ccccc|c|c}
\toprule
\multirow{2}{*}{\textbf{Method}} & \multicolumn{2}{c|}{\textbf{ScanMatch} $\uparrow$} & \multicolumn{5}{c|}{\textbf{MultiMatch} $\uparrow$}        & \multirow{2}{*}{\textbf{SED} $\downarrow$} & \multirow{2}{*}{\textbf{STDE} $\uparrow$} \\
                        \cline{2-3} \cline{4-8}  
                        & w/o Dur.             & w/ Dur.            & Vector & Direction & Length & Position & Duration &                                   &                                  \\ 
                        
                        \hline
IRL~\cite{zhibo:2020:cocosearch}                     & 0.1495               & -                  & 0.8248 & 0.6402    & 0.7688 & 0.6998   & -   & 6.6250                             & 0.7043                           \\
FFMs~\cite{zhibo:2022:targetabsent}                     & 0.2766               & -                  & 0.8914 & 0.6567    & 0.8785 & 0.8140    & -   & 5.9221                            & 0.8055                           \\
ChenLSTM~\cite{xianyu:2021:vqa}                & 0.2751               & 0.2153             & 0.8825 & 0.6222    & 0.8731 & 0.7940    & 0.6384   & 5.3468                            & 0.7841                           \\
ChenLSTM-ISP~\cite{chen2024isp}            & 0.2863               & 0.2205             & 0.8847 & 0.6430     & 0.8721 & 0.7980    & 0.6504   & 5.2895                            & 0.7865                           \\
Gazeformer~\cite{sounak:2023:gazeformer}              & 0.2971               & 0.2042             & 0.9080  & 0.6506    & 0.9035 & 0.8147   & 0.5901   & 5.1024                            & 0.8030                            \\
Gazeformer-ISP~\cite{chen2024isp}           & 0.2736               & 0.2146             & 0.9038 & 0.6181    & 0.8892 & 0.8031   & 0.6755   & 5.1905                            & 0.7875                           \\
HAT~\cite{yang2024hat}                     & 0.3120                & -                  & 0.9064 & 0.6443    & 0.9065 & 0.8138   & -   & 5.0613                            & 0.8006                           \\ \hline
\textbf{Our \method}                  & \textbf{0.3321}               & \textbf{0.2232}             & \textbf{0.9173} & \textbf{0.6790}     & \textbf{0.9174} & \textbf{0.8293}   & \textbf{0.6951}   & \textbf{4.8831}                            & \textbf{0.8089}                           \\ \bottomrule
\end{tabular}
}

\label{table:sota-scanpath}
\end{table*}
%-------------------------------------------------------------------------

\begin{figure*}[!t]
    \centering
    \includegraphics[width=0.9\linewidth]{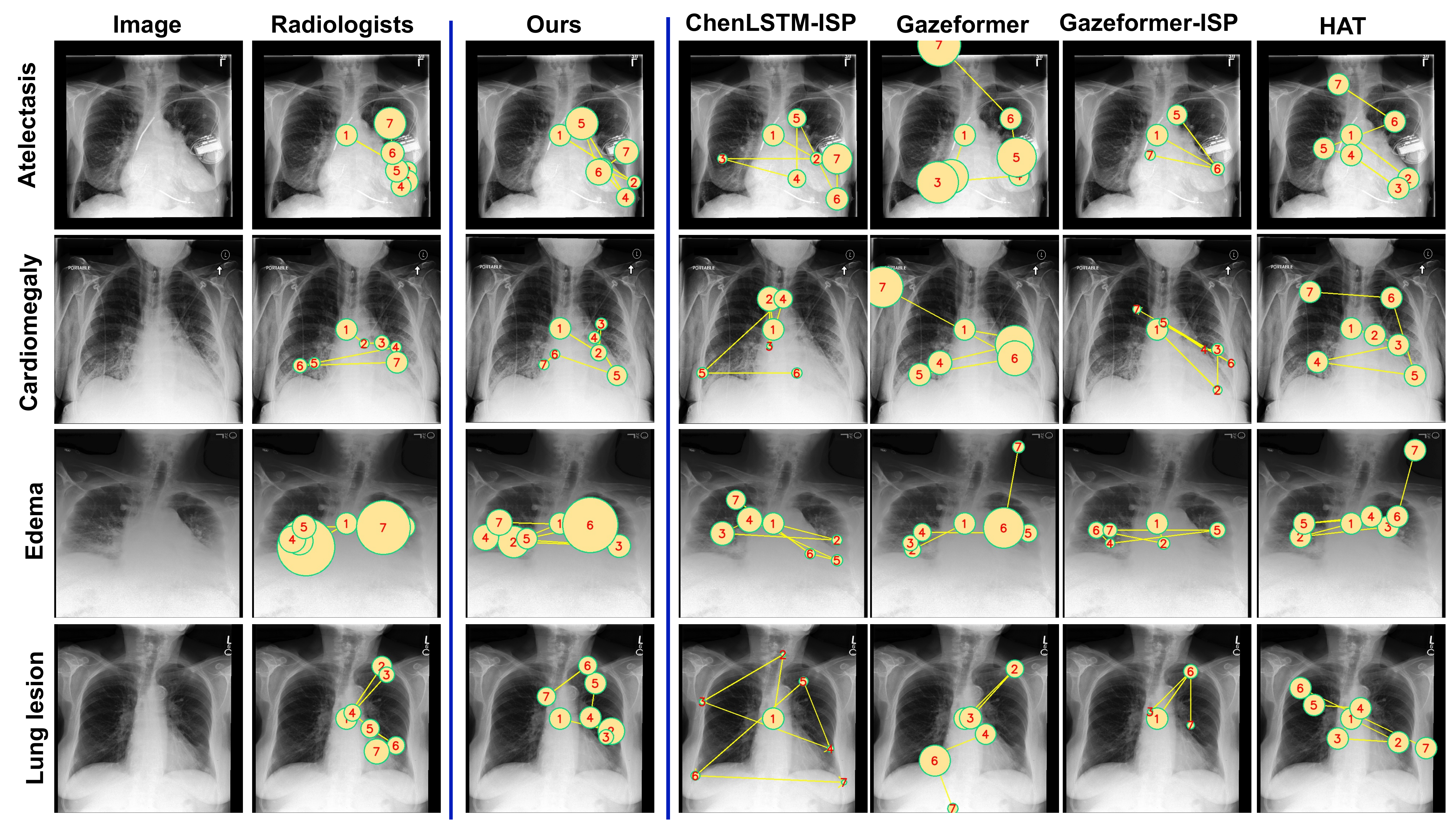}
    \caption{Qualitative results between our \method compared with ChenLSTM-ISP, Gazeformer, Gazeformer-ISP, and HAT. Four different findings (rows) including Atelectasis, Cardiomegaly, Edema, and Lung lesion are shown from the top to bottom. Each circle represents a fixation, with the number and radius indicating its order and duration, respectively. As HAT only predicts 2D coordinates, we let all predicted fixations of HAT have the same radius. }
    \vspace{-1em}
    \label{fig:qualitative-results}
\end{figure*}
%-------------------------------------------------------------------------

%-------------------------------------------------------------------------
% ABLATION START HERE

% ABLATION END HERE
%-------------------------------------------------------------------------

\section{Experiments}
\subsection{Implementation and Metrics}

\noindent
\textbf{Implementation details.} 
All images are scaled down to $224 \times 224$ for computing efficiency. 
The Fixation Decoder has $ L = 6$ layers with a hidden dimension $D = 384$. 
The MLP of Fixation Distribution Head consists of 384 units with 3 layers and ReLU activation.
\cref{eq:focal-pixel-loss} has $\alpha = 4$ $\gamma = 2$ based on the best validation results.
The Feature Extractor's backbone is ResNet-50~\cite{he2015resnet}, and we obtain the ResNet-50 checkpoint using MGCA~\cite{wang2022mgca} for 50 epochs with a batch size of 144. We then finetune this checkpoint jointly with the full pipeline.
We train the full pipeline for 30,000 iterations with a learning rate of \( 1 \times 10^{-5} \) and a batch size of 32.
The entire training process was conducted using AdamW~\cite{loshchilov2017adamw}, on a single A6000 GPU with 48GB of RAM.

\noindent
\textbf{Metrics.}
We evaluate fixation scanpath prediction accuracy using various metrics: ScanMatch~\cite{cristino2010scanmatch,sogo2013gazeparser} applies the Needleman-Wunsch algorithm~\cite{needleman1970general} to compare fixation locations and durations; MultiMatch~\cite{dewhurst2012depends} assesses similarity across five dimensions; String-Edit Distance (SED)~\cite{brandt1997spontaneous,foulsham2008can} compares character strings representing image regions; and Scaled Time-Delay Embedding (STDE)~\cite{wei:2011:stde} measures mean minimum Euclidean distances between sub-sequences of predicted and ground truth scanpaths.

\noindent
\textbf{Compared Methods.} We evaluate several state-of-the-art (SOTA) visual search methods on our \dataset: IRL~\cite{zhibo:2020:cocosearch}, FFMs~\cite{zhibo:2022:targetabsent}, ChenLSTM~\cite{xianyu:2021:vqa}, Gazeformer~\cite{sounak:2023:gazeformer}, ChenLSTM-ISP~\cite{chen2024isp}, Gazeformer-ISP~\cite{chen2024isp}, and HAT~\cite{yang2024hat}. Note that Gazeformer and Gazeformer-ISP require a pretrained CLIP component to encode the finding names, so we replace its default CLIP with BiomedCLIP~\cite{zhang2023biomedclip}. We adhere to the original training practices for all baselines. For more details, please refer to the Supplementary.

\subsection{Quantitative results}

\cref{table:sota-scanpath} demonstrates the proposed method's superior performance, surpassing SOTA approaches. Note that IRL, FFMs, and HAT do not predict fixation duration, so their evaluation on this metric is excluded. IRL and FFMs face challenges with sample efficiency due to reinforcement learning pipelines, while ChenLSTM variants and ISP methods are limited by their specialized modules—ChenLSTM relies on pretrained object detectors and ISPs on Observer-Centric modules.
HAT and Gazeformer overgeneralize and fail to fully leverage domain-specific information by design, with HAT ignoring duration data and Gazeformer relying heavily on CLIP for zero-shot visual search.
Our method avoids these limitations. High scores in metrics such as ScanMatch, MultiMatch, SED, and STDE demonstrate our method's capability to effectively capture complex scanpath dynamics, setting a new standard in chest X-ray target-present visual search.

\subsection{Qualitative results}

\cref{fig:qualitative-results} presents a qualitative comparison of scanpath patterns across different radiology findings and models, including radiologists and several state-of-the-art approaches. Generally, \method predicts more consistent and radiologist-like fixations than other methods.
ChenLSTM-ISP often exhibits scattered, less focused patterns, while Gazeformer-ISP may overlook key areas or focus on fewer locations. Although Gazeformer aligns better with ground truth than its ISP variant, it occasionally misses critical regions, such as lung lesions. HAT performs reasonably well but frequently covers the entire lung, even when attention should be limited to smaller areas, such as in cardiomegaly. In contrast, our \method shows fixation patterns more closely resembling those of radiologists, outperforming other state-of-the-art methods. Overall, \cref{fig:qualitative-results} underscores the effectiveness of our approach in mimicking expert gaze patterns across different findings. Additional comparison will be included in the Supplementary.

\subsection{Ablation study}
\label{sec:ablation-study}

To study the design choice of our proposed architecture, we ablate our method under several aspects. 
% More ablation studies are provided in the Supplementary Materials. All experiments are performed on our dataset, with the same training settings. Here, we report the average of all aspects for the MultiMatch score 

\noindent
\textbf{The importance of low- and high-resolution feature maps.} In \cref{sec:embedding}, guided by our intuition, we use two feature maps: a low-resolution map for high-level visual understanding and a high-resolution map for detailed visual understanding. These are concatenated into a single tensor for the Self-Attention layer, with the low-resolution feature serving as a \textit{reference} and the high-resolution feature \textit{indexed} using 2D Spatial Indexing to generate temporal features.
Ablation results in \cref{table:ablation_low_high_input} show that omitting 2D Spatial Indexing results in a significant performance drop due to the loss of temporal information. Conversely, not using the reference feature before Self-Attention has a lesser impact. The optimal performance is achieved by using the low-resolution feature as the reference and the high-resolution feature for 2D indexing, aligning with our intuitive design choices.
\begin{table}[!t]
\centering
\caption{The role of low- and high-resolution feature maps.} 
\resizebox{\linewidth}{!}{
\begin{tabular}{cc|cc|c|c|c}
\toprule
\multirow{2}{*}{\textbf{Reference} } &   \multirow{2}{*}{\textbf{Indexing} }               & \multicolumn{2}{c|}{\textbf{ScanMatch} $\uparrow$} & \multirow{2}{*}{\textbf{MultiMatch} $\uparrow$} & \multirow{2}{*}{\textbf{SED} $\downarrow$} & \multirow{2}{*}{\textbf{STDE} $\uparrow$} \\ \cline{3-4}
            &              & w/o Dur.            & w/ Dur.             &                                        &                                   &                                  \\ \hline
$P^l$                 & \xmark & 0.1848              & 0.2029              & 0.7070                                 & 6.3636                            & 0.7066                           \\
$P^h$                 & \xmark & 0.1939              & 0.1925              & 0.7058                                 & 6.1424                            & 0.7184                           \\ \hline
\xmark & $P^l$                 & 0.3077              & 0.2177              & 0.7927                                 & 5.0180                            & 0.8027                           \\
\xmark & $P^h$                 & 0.3176              & 0.2204              & 0.7985                                 & 4.9078                            & 0.8035                           \\ \hline
$P^l$                 & $P^l$                 & 0.3129              & 0.2228              & 0.7989                                 & 4.9100                            & 0.8060                           \\
$P^h$                 & $P^h$                 & 0.3221              & 0.2229              & 0.8015                                 & 5.0277                            & 0.8058                           \\
$P^h$                 & $P^l$                 & 0.3184              & 0.2210              & 0.8022                                 & 5.0224                            & 0.8057                           \\ \hline
$P^l$                 & $P^h$                 & \textbf{0.3321}     & \textbf{0.2232}     & \textbf{0.8076}                        & \textbf{4.8831}                   & \textbf{0.8089}                  \\ \bottomrule
\end{tabular}%
}
\vspace{-0.5em}
\label{table:ablation_low_high_input}
\end{table}

\noindent
\textbf{Initial Feature Extractor's weight contribution.} This ablation studies the effect of the initial weight for the Feature Extractor(\cref{sec:feature_extractor}), shown in \cref{table:ablation_weight_init}. In conclusion, using ImageNet checkpoint can give a decent performance. But with a better checkpoint, the performance is higher. This shows the robustness of our architecture.
%-------------------------------------------------------------------------
\begin{table}[!t]
\centering
\caption{Ablation study of choosing initial weight.} 
\resizebox{\linewidth}{!}{%
\begin{tabular}{l|cc|c|c|c}
\toprule
\multirow{2}{*}{\textbf{Inital Weight}} & \multicolumn{2}{c|}{\textbf{ScanMatch} $\uparrow$} & \multirow{2}{*}{\textbf{MultiMatch} $\uparrow$} & \multirow{2}{*}{\textbf{SED} $\downarrow$} & \multirow{2}{*}{\textbf{STDE} $\uparrow$} \\ \cline{2-3}
                        & w/o Dur.            & w/ Dur.             &                                        &                                   &                                  \\ \hline
Random Init.            & 0.3130              & 0.2205              & 0.79224                                & 5.0331                            & 0.8058                           \\
ImageNet                & 0.3238              & 0.2224              & 0.79942                                & 4.9723                            & 0.8081                           \\ \hline
Ours (Self-supervised)        & \textbf{0.3321}     & \textbf{0.2232}     & \textbf{0.80762}                       & \textbf{4.8831}                   & \textbf{0.8089}                  \\ \bottomrule
\end{tabular}%
}
\vspace{-1em}
\label{table:ablation_weight_init}
\end{table}

\section{Conclusion}
This paper addresses two key challenges: ambiguous fixations in existing eye-tracking datasets and the absence of a finding-aware radiologist’s scanpath model. Drawing inspiration from visual search datasets in general domains, we align findings with fixations, manage fixation durations using a radius-based heuristic, and constrain fixations on duration to produce the first finding-aware visual search dataset, \dataset. Our dataset reflects two key properties of visual search behavior: \#1 late fixations tend to converge on decisive regions of interest, and \#2 fixations within objects of interest are typically longer in duration compared to those outside.
We then propose \method that utilizes self-supervised training to obtain a medical pretrained feature extractor and a query mechanism to select relevant fixations for predicting subsequent ones. The extensive benchmark shows \method’s ability to generate radiologist-like scanpaths, serving as a strong baseline for future research.

\textbf{Discussion:} Our work impacts the behavioral vision literature in the medical domain, where (i) modeling and replicating radiologists' behavior has not been explored, (ii) understanding the understanding of finding-aware visual search and their integration with Deep Learning remains poorly understood~\cite{neves2024shedding}. These are critical for advancing diagnostics in radiology, enhancing decision-making processes, and enabling the future development of collaborative interactions between radiologists and AI systems.
% More broadly, our benchmark serves as the first dataset specifically designed for medical finding search prediction, setting the foundation for research across a wide range of medical applications, including human-computer interaction, error correction, workflow optimization, and expert analysis. Future work will focus on expanding the dataset to include target-absent samples, further enhancing the model's ability to handle complex diagnostic scenarios by identifying both the presence and absence of findings.

\noindent
\textbf{Acknowledgments.} This material is based upon work supported by the National Science Foundation (NSF) under Award No OIA-1946391, NSF 2223793 EFRI BRAID, National Institutes of Health (NIH) 1R01CA277739-01.

\clearpage
%%%%%%%%% REFERENCES
{\small
\bibliographystyle{ieee_fullname}
\bibliography{egbib}
}

\end{document}